\documentclass[10pt,twocolumn,letterpaper]{article}

\usepackage{iccv}
\usepackage{times}
\usepackage{epsfig}
\usepackage{graphicx}
\usepackage{amsmath}
\usepackage{amssymb}
\usepackage{gensymb}
\usepackage{subfig}
\usepackage[pagebackref=true,breaklinks=true,letterpaper=true,colorlinks,bookmarks=false]{hyperref}

\iccvfinalcopy 

\def\iccvPaperID{10} 

\ificcvfinal\fi
\begin{document}

\title{SaltiNet: Scan-path Prediction on 360 Degree Images using Saliency Volumes}

\author{Marc Assens\thanks{Work developed while Marc Assens was a visiting student at Insight Center for Data Analytics.} \thinspace and Xavier Giro-i-Nieto\\
Image Processing Group\\
Universitat Politecnica de Catalunya (UPC)\\
Barcelona, Catalonia/Spain\\
{\tt\small xavier.giro@upc.edu}
\and
{Kevin McGuinness and Noel E. O'Connor}\\
Insight Center for Data Analytics\\
Dublin City University\\
Dublin, Ireland\\
{\tt\small kevin.mcguinness@insight-centre.org}
}

\maketitle

\begin{abstract}
We introduce SaltiNet, a deep neural network for scanpath prediction trained on 360-degree images. The model is based on a temporal-aware novel representation of saliency information named the saliency volume. The first part of the network consists of a model trained to generate saliency volumes, whose parameters are fit by back-propagation computed from a binary cross entropy (BCE) loss over downsampled versions of the saliency volumes. Sampling strategies over these volumes are used to generate scanpaths over the 360-degree images. Our experiments show the advantages of using saliency volumes, and how they can be used for related tasks. Our source code and trained models available at \url{https://github.com/massens/saliency-360salient-2017}.
\end{abstract}


\section{Motivation}
\label{sec:Motivation}

Visual saliency prediction is a field in computer vision that aims to estimate the areas of an image that attract the attention of humans. This information can provide important clues to human image understanding. The data collected for this purpose are fixation points in an image, produced by a human observer that explores the image for a few seconds, and are traditionally captured with eye-trackers \cite{wilming2017extensive}, mouse clicks \cite{jiang2015salicon} and webcams \cite{cvpr2016_Khosla}. The fixations are usually aggregated and represented with a saliency map, a single channel image obtained by convolving a Gaussian kernel with each fixation. The result is a gray-scale heatmap that represents the probability of each pixel in an image being fixated by a human, and it is usually used as a soft-attention guide for other computer vision tasks.

Traditionally, saliency maps have only described fixation information with respect to the spatial layout of an image.
This type of representations only encode the probability of each image pixel capture the visual attention of the user, but with no information regarding the order in which these pixels may be scanned or the duration of the fixation.
Recent studies have raised the need for a representation that is also temporal-aware \cite{bylinskii2016should}.
We address the temporal challenge for the particular case of 360\degree\ images, which contain the complete scene around the capture point and allow the viewer to choose the observation angle.
Predicting the pattern that humans follow in 360\degree\ images is a topic of special interest for VR/AR applications, as it facilitates an efficient encoding and rendering on the display devices.

\begin{figure}
\centering
\includegraphics[width=1.0\columnwidth]{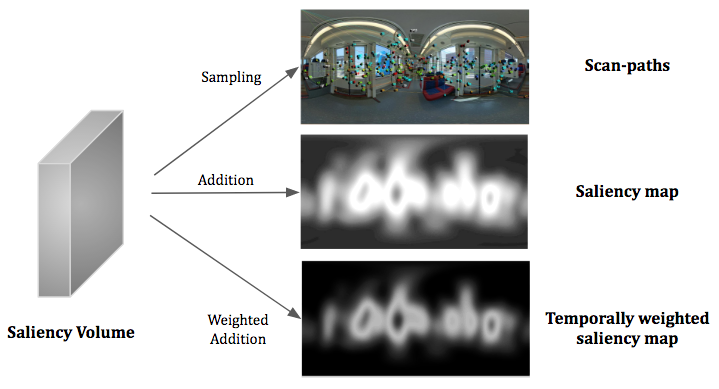}
\caption{Scan-paths, saliency maps and temporally weighted saliency maps can be generated from a saliency volume.}
\label{salvol}
\end{figure}

The main contributions of this paper are the following:

\begin{itemize}
\item the introduction of \textit{saliency volumes} to capture the temporal nature of eye-gaze scan-paths;
\item the SaltiNet architecture to generate scan-paths from a deep neural network that predicts saliency volumes and a sampling strategy over them;
\item this work has been awarded as the best scanpath solution at the Salient360! challenge from the IEEE International Conference on Multimedia and Expo (ICME) 2017 \cite{salient360}.
\end{itemize}


This paper is structured as follows. Section \ref{sec:related_work} reviews the related literature in saliency prediction for eye fixations and presents our work with respect to them. Section \ref{sec:Architecture} presents the whole architecture of the system, and Section \ref{sec:Training} describes how the deep neural network was trained. Section \ref{sec:Experiments} describes the experiments and results to assess the performance of the model, while Section \ref{sec:Conclusions} draws the conclusions and future work.





\begin{figure*}[!ht]
\includegraphics[width=\textwidth]{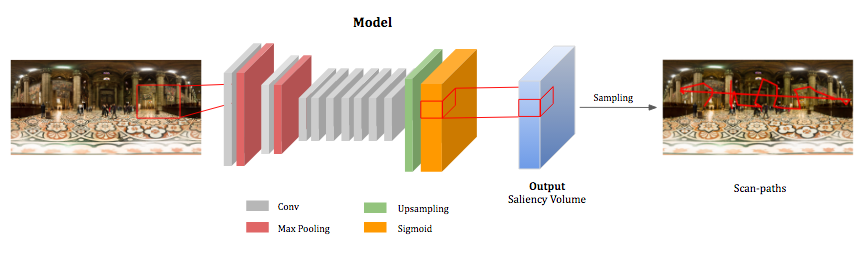}
\caption{Overall architecture of the proposed scanpath estimation system.}
\label{fig:architecture}
\end{figure*}

\section{Related Work}
\label{sec:related_work}

\subsection{Saliency prediction}
\label{subsec:saliency_prediction}

The first models for saliency prediction were biologically inspired and based on a bottom-up computational model that extracted low-level visual features such as intensity, color, orientation, texture and motion at multiple scales. Itti et al. \cite{itti1998model} proposed a model that combines multiscale low-level features to create a saliency map. Harel et al. \cite{harel2007graph} presented a graph-based alternative that starts from low-level feature maps and creates Markov chains over various image maps, treating the equilibrium distribution over map locations as activation and saliency values. 

Though this models did well qualitatively, the models had limited use because they frequently did not match actual human saccades from eye-tracking data. It seemed that humans not only base their attention on low-level features, but also on high-level semantics \cite{bylinskii2016should} (e.g., faces, humans, cars, etc.). Judd et al. introduced in \cite{judd2009learning} an approach that used low, mid and high-level image features to define salient locations. This features where used in combination with a linear support vector machine to train a saliency model. Borji \cite{borji2012boosting} also combined low-level features with top-down cognitive visual features and learned a direct mapping to eye fixations using Regression, SVM and AdaBoost calssifiers.

Recently, the field of saliency prediction has made great progress due to advance of deep learning and its applications on the task of image classification \cite{krizhevsky2012imagenet} \cite{zhou2014learning}. The advances suggest that these models are able to capture high-level features. As stated in \cite{bylinskii2016should}, in March of 2016 there where six deep learning models among the top 10 results in the MIT300 saliency Benchmark \cite{mit-saliency-benchmark}. 


The enormous amount of training data necessary to train these networks makes them difficult to train directly for saliency prediction. With the objective of allowing saliency models to capture this high-level features, some authors have adapted well-known models with good performance in the task of Image Recognition. DeepGaze \cite{kummerer2014deep} achieved state of the art performance by reusing the well-known AlexNet \cite{krizhevsky2012imagenet} pretrained on ImageNet \cite{deng2009imagenet} with a network on top that reads activations from the different layers of AlexNet. The output of the network is then blurred, center biased and converted to a probability distribution using a softmax. A second version called DeepGaze 2 \cite{kummerer2016deepgaze} used features from VGG-19 \cite{simonyan2014vgg} trained for image recognition. In this case, they did not fine-tune the network. Rather, some readout layers were trained on top of the VGG features to predict saliency with the SALICON dataset \cite{jiang2015salicon}. This results corroborated that deep features trained on object recognition provide a versatile feature space for performing related visual tasks.  
A complete new architecture designed and trained for saliency prediction was proposed in \cite{Pan_2016_CVPR}, but the same work also observed the benefits of using deeper pre-trained models for image classification as a basis.
Other advances in deep learning such as generative adversarial training (GANs) and attentive mechanisms have also been applied to saliency prediction: SalGAN \cite{pan2017salgan} is a deep network for saliency prediction that measured the gain in performance when using a universal adversarial training in opposite to optimizing for a specific loss function.
The Saliency Attentive Model (SAM) \cite{cornia2016predicting} includes a Convolutional LSTM that focuses on the most salient regions of the image to iteratively refine the predicted saliency map. 

In \cite{torralba2006contextual}, Torralba et al. studied how the scene modules visual attention and discovered that the same objects recieve different attention depending on the scene where they appear (i.e. pedestrians are the most salient object in only 10\% of the outdoor scene images, being less salient than many other objects. Tables and chairs are among the most salient objects in indoor scenes). With this insight, Liu et al. proposed DSCLRCN \cite{liu2016context}, a model based on CNNs that also incorporates global context and scene context using RNNs. Their experiments have obtained outstanding results in the MIT Saliency Benchmark.

Recently, there has been interest in finding appropiate loss functions. Huang et al. \cite{huang2015salicon} made an interesting contribution by introducing loss functions based on metrics that are differentiable, such as NSS, CC, SIM and KL divergence to train a network (see \cite{riche2013saliency} and \cite{kummerer2015metrics}).

\subsection{Scanpath prediction}
\label{subsec:scanpath_prediction}

Unlike with the related task of saliency map prediciton, there has not been much progress in the task of scanpath prediciton over the last years. Cerf et al. \cite{cerf2008predicting} discovered that observers, even when not instructed to look for anything particular, fixate on a human face with a probability of over 80\% within their first two fixations. Furthermore, they exhibit more similar scanpaths when faces are present. Recently, Hu et al. \cite{hu2017pilot} have introduced a model capable of selecting relevant areas of a 360\degree\ video and deciding in which direction should a human observer look at each frame. An object detector is used to propose candidate objects of interest and a RNN selects the main object at each frame. 


\section{Architecture}
\label{sec:Architecture}
\par The central element in the architecture of SaltiNet is a deep convolutional neural network (DCNN) that predicts a saliency volume for a given input image. This section provides detail on the structure of the network, the loss function, and the strategy used to generate scan-paths from saliency volumes.

\subsection{Saliency Volumes} 
\label{sub:saliency_volumes}

\par


Saliency volumes aim to be a suitable representation of spatial and temporal saliency information for images. They have three axes that represent the width and height of the image, and the temporal dimension. 

Saliency volumes are generated from information already available in current fixation datasets. First, the timestamps of the fixations are quantized. The length of the time axis is determined by the longest timestamp and the quantization step. Second, a binary volume is created by placing a `1' at fixation points and a `0' on the remaining positions. Third, a multivariate Gaussian kernel is convolved with the volume to generate the saliency volume. The values of each temporal slice are normalized, converting the slice into a probability map that represents the probability of each pixel being fixated by a user at each timestep. 

Figure \ref{salvol} shows how saliency volumes are a meta-representation of saliency information and how other saliency representations can be extracted from them. Saliency maps can be generated by performing an addition operation across all the temporal slices of the volume, and normalizing the values to ensure they add to one. A similar representation are \textit{temporally weighted saliency maps}, which are generated by performing a weighted addition operation of all the temporal slices. Finally, scan-paths can also be extracted by sampling fixation points from the temporal slices. Sampling strategies that aim to generate realistic scan-paths are will be discussed in Section~\ref{ssec:ContentLossExp}.


\subsection{Convolutional Neural Network}
\label{ssec:Neural network}

We propose a convolutional neural network (CNN) that adapts the filters learned to predict flat saliency maps to predict saliency volumes. Figure \ref{fig:architecture} illustrates the  architecture of the convolutional neural network, composed of 10 layers and a total of 25.8 million parameters. Each convolutional layer is followed by a rectified linear unit non-linearity (ReLU). Excluding the last layer, the architecture follows the proposal of SalNet \cite{Pan_2016_CVPR}, whose first three layers are initialized from the VGG-16 model \cite{chatfield2014return} trained for image classification. 

Our network was designed considering the amount of training data available. Different strategies where introduced to prevent overfitting. First, the model was previously trained on the similar task of saliency map prediction, and the obtained weights were fine-tunned for the task of saliency volume prediction. Second, the input images where resized to $[300 \times 600]$, a much smaller dimension than their initial size $[3000 \times 6000]$. The last layer of the network outputs a volume of size $[12 \times 300 \times 600]$, with three axis that represent time, height, and width of the image. 

\subsection{Scan-path sampling}
\label{ssec:sampling}

We take a stochastic approach to scan-path sampling\footnote{We also experimented with using an LSTM to directly predict scan-paths from the training data. However, we found that this resulted in the model regressing to the image center \cite{mathieu2015deep}. Future work will consider using adversarial training to address this.}. The generation of scan-paths from the saliency volumes requires determining: 1) number of fixations of each scan-path; 2) the duration in seconds of each fixation; and 3) the location of each fixation point. 
The first two values were sampled from their probability distributions learned from the training data.
The location of each fixation point was also generated by sampling, this time from the corresponding temporal slice from the predicted saliency volume.
Different strategies were explored for this purpose, presented together with their performance in Section \ref{sec:Experiments}.


\section{Training}
\label{sec:Training}


We trained the CNN in SaltiNet over 36 images of the 40 training images from the Salient360 dataset \cite{salient360}, leaving aside 4 images for validation. 
We normalized the values of the saliency volumes to be in the interval of [0, 1]. Both the input images and the saliency volumes were downsampled to $600 \times 300$ prior to training. The saliency volumes where generated from fixations using a multivariate Gaussian kernel with bandwidths $[4, 20, 20]$ (time, height, width).

The CNN was trained using stochastic gradient descent with cross entropy loss using a batch size of 1 image during 90 epoch. The binary cross entropy loss is defined as ${L}_{BCE}$ in Eq.~\eqref{eq:1}, where $S_j$ and $\hat{S}_j$ correspond to the ground truth and predicted values of the saliency map.

\begin{equation}\label{eq:1}
\begin{split}
\mathcal{L}_{BCE} = 
-\frac{1}{N}
\sum_{j=1}^{N} 
S_j \log(\hat{S}_j) + (1 - S_j)\log(1 - \hat{S}_j).  
\end{split}
\end{equation}

\begin{figure}
\centering
\includegraphics[width=1.0\columnwidth]{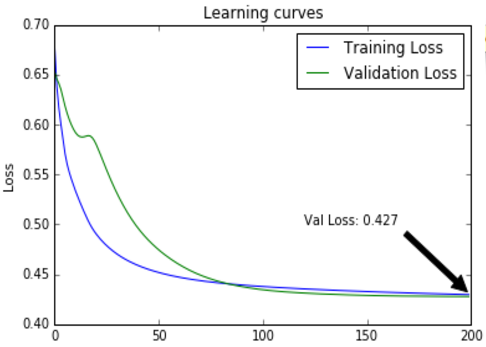}
\caption{Training curves for our model (binary cross entropy loss.)}
\label{fig:learning_curves}
\end{figure}

During training, results on the validation set were tracked to monitor convergence and overfitting problems. The $L^2$ weight regularizer (weight decay) was used to avoid overfitting. Our network took approximately two hours to train on a NVIDIA GTX Titan X GPU using the Keras framework with the Theano backend. The learning rate was set to $\alpha=0.001$ throughout training. Figure~\ref{fig:learning_curves} shows the learning curves.


\section{Experiments}
\label{sec:Experiments}

SaltiNet was assessed and compared from different perspectives. First, we assess the impact of different sampling strategies to generate scan-paths from saliency volumes. Second, we show quantitative performance results of the model. 

\subsection{Datasets}

Due to the small size of the training dataset, we performed transfer learning to initialize the weights of the network using related tasks. First, the network was trained to predict saliency maps using the SALICON dataset \cite{huang2015salicon} using the same architecture of SalNet \cite{Pan_2016_CVPR}. Then, the network was trained to predict saliency volumes generated from the iSUN dataset \cite{xu2015turkergaze} that contains 6000 training images. 
The network was fine-tuned using the 60 images of the dataset of head and eye movements provided by the University of Nantes \cite{rai2017dataset}.
This dataset was acquired based on the images displayed on the head mounted display (HMD) Oculus-DK2.
Eye gaze data was captured from a Sensomotoric Instruments (SMI) sensor in the HMD, which transmitted eye-tracking data binocularly at 60Hz.
There were 40-42 observers, who could freely observe the scene with no task instructed. Each 360 images were shown for 25 seconds and there was a 5 second gray screen between two images.

\subsection{Metric}

The similarity metric used is a variation of the Jarodzka algorithm \cite{jarodzka2010vector} proposed by the authors of the 360 saliency dataset \cite{rai2017dataset}. The standard similarity criteria was slightly modified to use equirectangular distances in 360 instead of Euclidean distances. The generated and ground truth scan-paths are matched 1 to 1 using the Hungarian algorithm to obtain the minimum possible final cost.
The presented results compare the similarity of 40 generated scan-paths with the scan-paths in the ground truth.




\subsection{Sampling strategies}
\label{ssec:ContentLossExp}

\begin{figure}
\centering
\includegraphics[width=1.0\columnwidth]{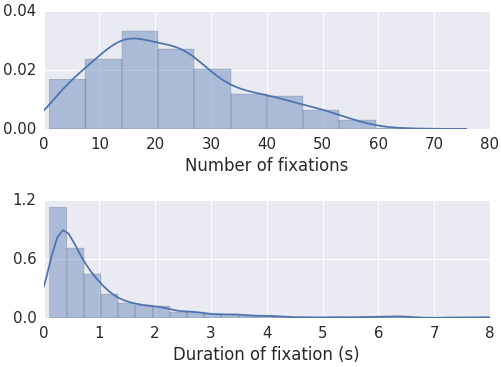}
\caption{Empirical distributions of the number of fixations per scan-paths (top) and duration of each fixation (bottom).}
\label{fig:distributions}
\end{figure}

Figure \ref{fig:distributions} shows the distribution of the number of fixations and the duration of each fixations for the training set. During scan path generation, we sample the number of fixations and their duration from these empirical distributions.

Regarding the spatial location of the fixation points, three different strategies were explored. 
The simplest approach (1) consists of taking one fixation for each temporal slice of the saliency volume. Through qualitative observation we noticed that scan-paths generated in this way were unrealistic, as the probability of each fixation is not conditioned on previous fixations. A more elaborated sampling strategy (2) consists of forcing fixations to be closer to their respective previous fixation. This is accomplished by multiplying a temporal slice (probability map) of the saliency volume with a Gaussian kernel centered at the previous fixation point. This suppresses the probability of positions that are far from the previous fixation point. The third sampling strategy (3) we assessed consisted of suppressing the area around all previous fixations using Gaussian kernels. As shown in Table \ref{tab:results}, we found that the best performing model was the one using sampling strategy (2).

\subsection{Results} 
\label{sub:results}
Scan-path prediction evaluation has received attention lately and it is a very active field of research \cite{le2013methods}\cite{jarodzka2010vector}.

Table \ref{tab:results} presents the impact of different sampling strategies over the saliency volume. We have compared our results with the accuracy that would be obtained by a model that outputs random fixations, and a model that outputs the ground truth fixations.



\begin{table}
\begin{center}
\begin{tabular}{lc}
\hline
			& Jarodzka$\downarrow$ \\
\hline
Random scan-paths & 4.94\\

\hline
(1) Naive sampling strategy & 3.45\\
(3) Avoiding fixating on same places & 2.82\\
(2) Limiting distance between fixations & \textbf{2.27}\\
\hline
Sampling ground truth saliency map & 1.89\\
Sampling ground truth saliency volume & 1.79\\
Ground truth scan-paths & 1.2e-8\\
\hline
\end{tabular}
\end{center}
\caption{Comparison between the three considered spatial sampling strategies. Lower values are better. }
\label{tab:results}
\end{table}

Table \ref{tab:icme} compares our system with two other solutions presented at the Salient360! Challenge \cite{salient360} held at the 2017 IEEE ICME conference in Hong Kong.
These figures were provided by the organizers of the challenge.
Results clearly indicate the superior performance of our system with respect to the two other participants.

\begin{figure}[h!]
\subfloat[Example of predicted scan-path]{\includegraphics[clip,width=1.0\columnwidth]{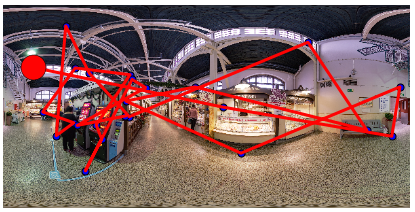}}
\par
\subfloat[Example of ground truth scan-path]{\includegraphics[clip,width=1.0\columnwidth]{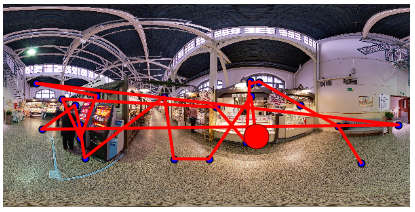}}
\caption{The top image shows a predicted scanpath, sampled from a predicted saliency volume. The image at the bottom shows a single ground truth scanpath.}
\label{fig:examples_scanpaths}
\end{figure}

\begin{figure*}[htp]
\subfloat[Indoor example]{\includegraphics[clip,width=0.9\linewidth]{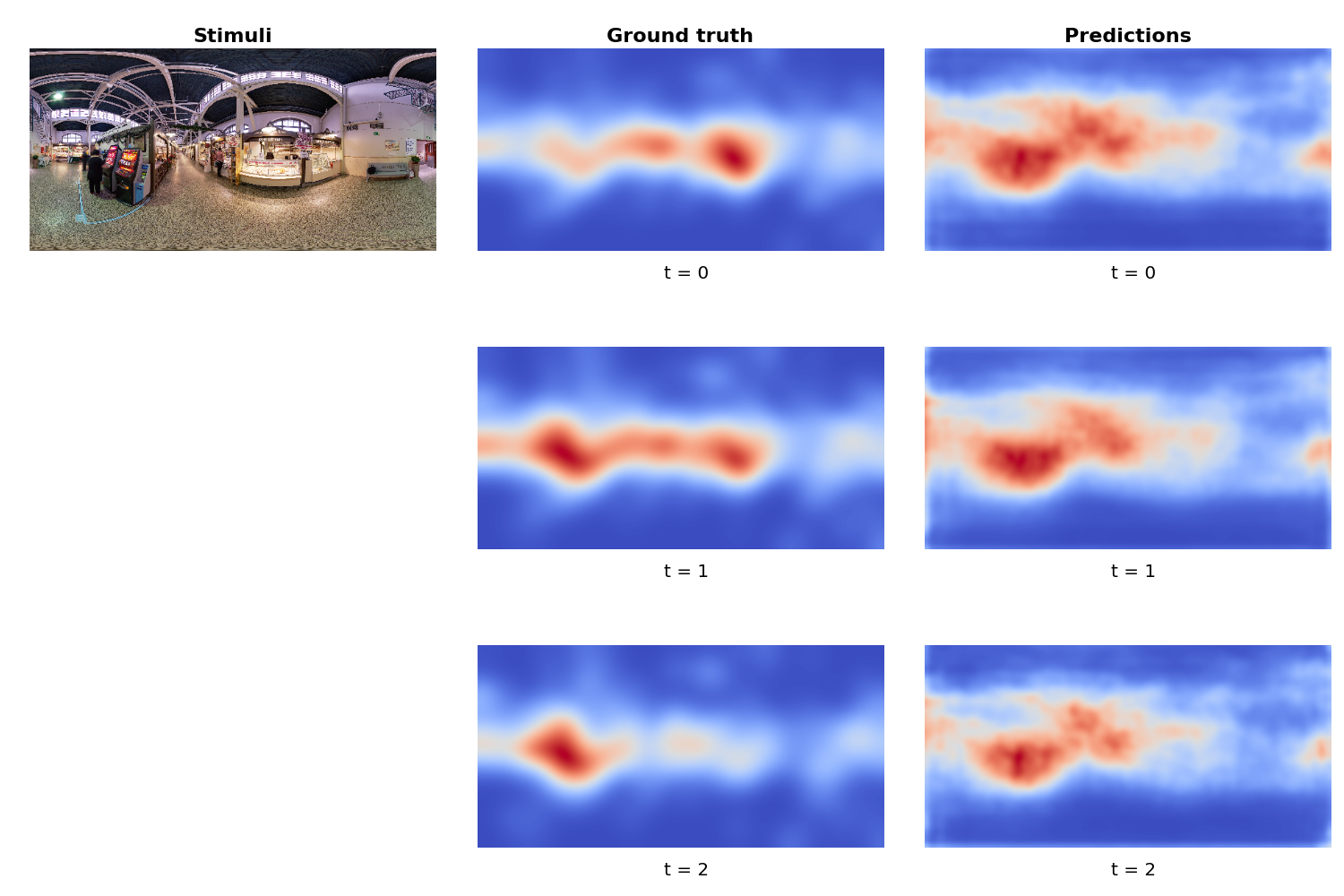}}
\quad
\subfloat[Outdoor example]{\includegraphics[clip,width=0.9\linewidth]{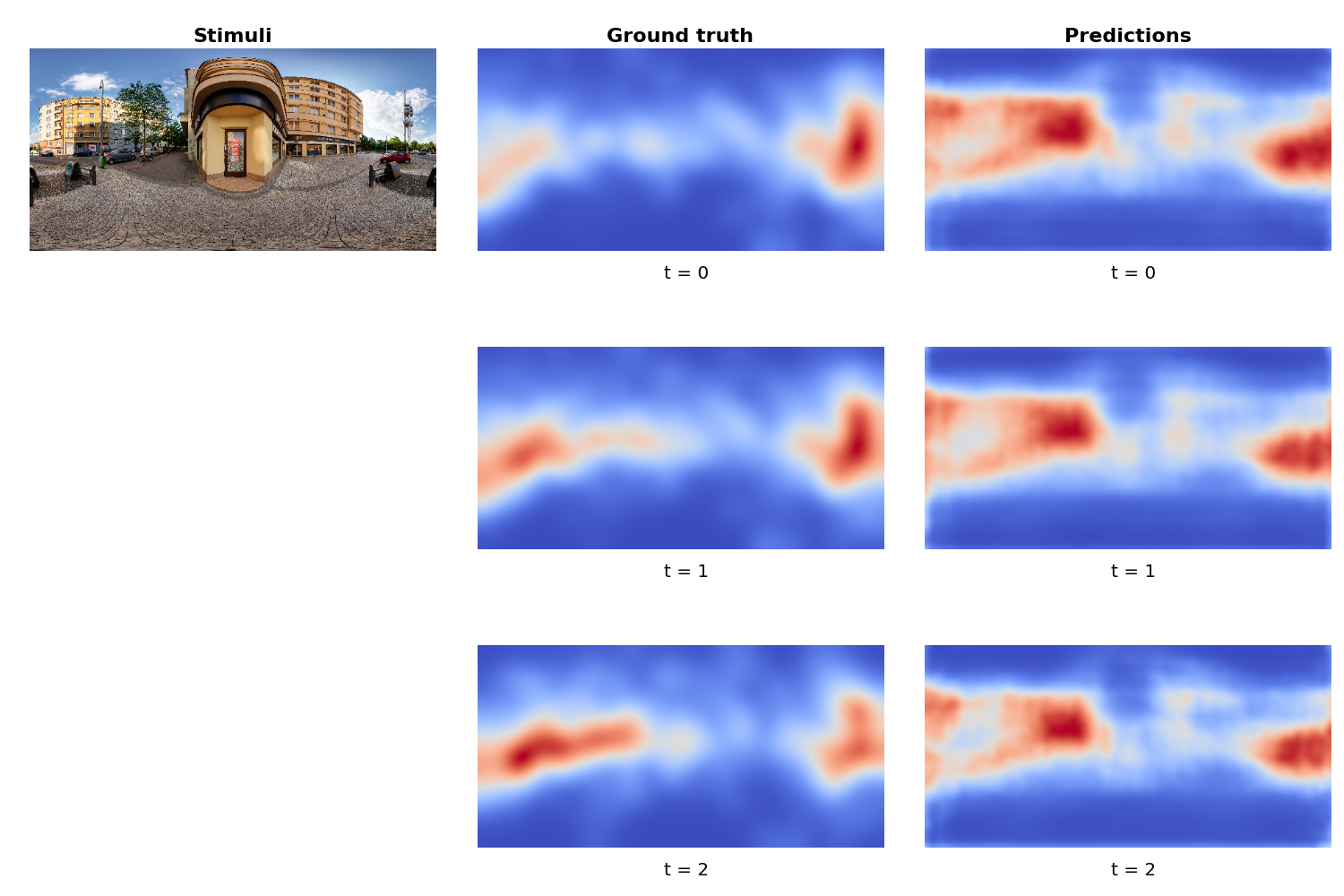}}

\caption{The images above show the predicted and ground truth saliency volumes for a given stimulus. For each saliency volume, three temporal slices are shown.}
\label{fig:examples_salvols}
\end{figure*}

The performance of our model has also been explored from a qualitative perspective by observing the generated saliency volumes and scan-paths. Figure \ref{fig:examples_scanpaths} compares a generated scan-path with a ground truth scan-path. Figure \ref{fig:examples_salvols} shows two examples of ground truth and generated saliency volumes. 

\begin{table}
\begin{center}
\begin{tabular}{lc}
\hline
							& Jarodzka$\downarrow$ \\

\hline
\textbf{SaltiNet} (Ours)			& \textbf{2.8697}		\\
SJTU 						& 4.6565\\
Wuhan University			& 5.9517\\
\hline
\end{tabular}
\end{center}
\caption{Comparison between three submissions to the Salient360! Challenge. Lower values are better. }
\label{tab:icme}
\end{table}



\section{Conclusions}
\label{sec:Conclusions}

In this work we have presented SaltiNet, a model capable of predicting scan-paths on 360\degree\ images. We have also introduced a novel temporal-aware saliency representation that is able to generate other standard representations such as scanpaths, saliency maps or temporally weighted saliency maps. Our experiments show that it is possible to obtain realistic scanpaths by sampling from saliency volumes, and the accuracy greatly depends on the sampling strategy. 



We have also found the following limitations to the generation of scanpaths from saliency volumes: 1) the probability of a fixation is not conditioned to previous fixations; 2) the length of the scanpaths and the duration of each fixation are treated as independent random variables. We have tried to address the first problem by using more complex sampling strategies. Nevertheless, this three parameters are not independently distributed and therefore our model is not able to accurately represent this relationship. 
Future work will aim at training a fully end-to-end  neural network capable of prediction the scan-paths with no need of the sampling module. 

\par Our results can be reproduced with the source code
and trained models available at \url{https://github.com/massens/saliency-360salient-2017}.


\section{Acknowledgments}
The Image Processing Group at UPC is supported by the project TEC2013-43935-R and TEC2016-75976-R, funded by the Spanish Ministerio de Economia y Competitividad and the European Regional Development Fund (ERDF). The Image Processing Group at UPC is a SGR14 Consolidated Research Group recognized and sponsored by the Catalan Government (Generalitat de Catalunya) through its AGAUR office. 
This material is based upon works supported by Science Foundation Ireland under Grant No 15/SIRG/3283.
We gratefully acknowledge the support of NVIDIA Corporation for the donation of GPUs used in this work.

{\small
\bibliographystyle{ieee}
\bibliography{egbib}
}

\end{document}